\documentclass[conference]{IEEEtran}
\IEEEoverridecommandlockouts
\usepackage{cite}
\usepackage{amsmath,amssymb,amsfonts}
\usepackage{algorithmic}
\usepackage{graphicx}
\usepackage{balance}
\usepackage{textcomp}
\usepackage{xcolor}
\usepackage{url}

\IEEEpubid{}

\def\BibTeX{{\rm B\kern-.05em{\sc i\kern-.025em b}\kern-.08em
    T\kern-.1667em\lower.7ex\hbox{E}\kern-.125emX}}
\begin{document}

\title{FireNet: A Specialized Lightweight Fire \& Smoke Detection Model for Real-Time IoT Applications}

\makeatletter
\newcommand{\linebreakand}{%
  \end{@IEEEauthorhalign}
  \hfill\mbox{}\par
  \mbox{}\hfill\begin{@IEEEauthorhalign}
}
\makeatletter

\author{\IEEEauthorblockN{Arpit Jadon, \textit{Student Member, IEEE}}
\IEEEauthorblockA{\textit{Z.H. College of Engg. and Tech.} \\
\textit{Aligarh Muslim University}\\
Aligarh, India \\
arpitjadon@zhcet.ac.in}
\and

\IEEEauthorblockN{Mohd. Omama}
\IEEEauthorblockA{\textit{Z.H. College of Engg. and Tech.} \\
\textit{Aligarh Muslim University}\\
Aligarh, India \\
mohd.omama@gmail.com}
\and

\IEEEauthorblockN{Akshay Varshney}
\IEEEauthorblockA{\textit{Z.H. College of Engg. and Tech.} \\
\textit{Aligarh Muslim University}\\
Aligarh, India \\
akshayvarshney.001@gmail.com}

\linebreakand 

\IEEEauthorblockN{Mohammad Samar Ansari, \textit{Member, IEEE}}
\IEEEauthorblockA{\textit{Software Research Institute} \\
\textit{Athlone Institute of Technology}\\
Ireland \\
mansari@ait.ie}
\and

\IEEEauthorblockN{Rishabh Sharma}
\IEEEauthorblockA{\textit{Z.H. College of Engg. and Tech.} \\
\textit{Aligarh Muslim University}\\
Aligarh, India \\
rishabh.sharma079@gmail.com}
}

\maketitle

\begin{abstract}
Fire disasters typically result in lot of loss to life and property. It is therefore imperative that precise, fast, and possibly portable solutions to detect fire be made readily available to the masses at reasonable prices. There have been several research attempts to design effective and appropriately priced fire detection systems with varying degrees of success. However, most of them  demonstrate a trade-off between performance and model size (which decides the model's ability to be installed on portable devices). The work presented in this paper is an attempt to deal with both the performance and model size issues in one design. Toward that end, a `designed-from-scratch' neural network, named \textit{FireNet}, is proposed which is worthy on both the counts: (i) it has better performance than existing counterparts, and (ii) it is lightweight enough to be deploy-able on embedded platforms like Raspberry Pi. Performance evaluations on a standard dataset, as well as our own newly introduced custom-compiled fire dataset, are extremely encouraging.
\end{abstract}


\section{Introduction}
Due to the rapid increase in the number of fire accidents, the fire alarm systems form an integral part of the necessary accessories in any sort of construction. Fire accidents are the most commonly occurring disasters nowadays. In order to mitigate the number of fire accidents, a large number of methods have been proposed for early fire detection to reduce the damage caused by such accidents. Apart from the problem of early fire detection, present fire alarm systems also prove to be inefficient in terms of the false triggering of the alarm systems. Present fire detection methods are typically based on physical sensors like thermal detectors, smoke detectors, and flame detectors. However, these sensor-based detection systems are not very reliable for fire detection. For instance, quite often there are cases of false triggering with smoke detectors, as it does not possess the capability to differentiate between fire and smoke. On the other hand, the other two detection systems need a sufficient level of fire initiation for a clear detection, which  leads to a long detection delay causing irreparable damages. An alternative, which could lead to the enhancement of robustness and reliability in the present fire detection systems, is the visual fire detection approach.

As a courtesy of the advancement in various artificial intelligence fields, vision-based research fields like Image Processing and Computer Vision have witnessed a fair share of fruitful benefits. Various deep learning (DL) \cite{lecun2015deep} models have been able to comfortably surpass the human level performance in specific computer vision applications like image classification. The visual-based fire detection approach also has been able to take advantage of these technological improvements. Visual-based fire detection systems have many advantages over the hardware-based alarm systems in terms of cost, accuracy, robustness, and reliability. Over time the handcrafted visual fire detection approaches, which offer lower performance in terms of accuracy and false triggering, have been replaced by deep learning based approaches, which are better in performance in terms of varying metrics. This better performance is attributed  to the capability of the deep learning based approaches to automatically extract features from the raw images. On the contrary, the handcrafted approaches require more careful handling as the features are to be extracted manually from the input images. Thus, by combining these more reliable visual based fire detection techniques with the conventional sensor based techniques, a more robust and reliable fire alarm system could be developed. Therefore, in this work, our aim is to introduce an advanced fire and smoke detection unit, which is reliable, reduces the false triggering problem and incorporates various state of the art technological concepts like Convolutional Neural Networks (CNN) and the Internet of Things (IoT) \cite{gubbi2013internet}. In order to get the best fire detection performance while maintaining a significantly good frame rate on the Raspberry Pi 3B (1.2GHz Broadcom BCM2837 64bit CPU, 1 GB RAM computer \cite{color6}) during real-time fire detection, we have developed our own neural network from scratch and have trained it on a dataset compiled from multiple sources.  The model is tested on a real world fire dataset, accumulated by us and also on a standard fire dataset provided by authors in the work \cite{foggia2015real}. We have taken Raspberry Pi as the hardware platform to deploy our model because it is the most cost-effective platform for running computationally non-intensive deep learning algorithms and will serve well our purpose of developing a fire and smoke detection unit. In addition, the system is also capable of differentiating fire from smoke thereby reducing the false triggering problem by triggering distinct sound  alarms for fire and smoke respectively. 

However, smoke detection using vision-based techniques face many challenges. Video processing techniques generally work on the principle of reading pixel values of the color. Thus, in real time, it becomes difficult to distinguish between the greyish colored objects present in the image along with smoke to be detected. Moreover, smoke detection becomes harder in a dark environment due to the camera's capturing attributes \cite{chen2006smoke}. Hence, for better overall performance, in our work, we utilize a smoke sensor integrated with the system for smoke detection, thus also removing the need for installing separate fire and smoke detectors. It simultaneously minimizes the false triggering problem occurring with smoke detectors used as fire detectors in conventional fire alarm systems. Smoke sensors are also economically affordable and can detect smoke efficiently. To alert the end user about the fire emergency, an IoT based remote data transmission system is also employed to send MMS containing visual fire feedback and fire alert to the end user.

This kind of intelligent fire and smoke detection unit can be used for a wide range of applications, such as giving an early warning for an emergency, notifying fire brigades so that they can get to the site of fire as soon as possible, triggering the automatic fire suppression systems, etc. \cite{ColorDifference}.

\subsection{Contributions}

The main contributions of this work are as follows:

\begin{itemize}

\item We introduce a shallow neural network for fire detection, which unlike previous DL-based fire detection approaches where bulky convolutional neural networks were used, can perform real-time fire detection at a frame rate that surpasses the frame rates achieved until now.

\item  We also introduce a new, small but very diverse,  training fire dataset combining images from multiple sources. Moreover, we also introduce another self-made dataset consisting of self-shot videos in a challenging environment. 

\item We also present a working implementation of a complete fire detection unit that can suitably replace the conventional physical sensor based fire detection alarm systems, simultaneously reducing the false and delayed triggering problems, along with providing a remote verification functionality by providing real-time visual feedback in the form of an alert message using Internet of Things (IoT).

\end{itemize}

This paper is organized as follows. Section  II discusses the past research work done in the area of deep learning and hand-engineering based fire detection. Section  III provides a detailed description of our dataset followed by the description of our proposed approach in section IV, which is in turn followed by the section  V, which discusses the IoT implementation to develop a complete state of the art fire detection unit. Finally, we discuss our results in section VI, followed by a discussion on the effectiveness of the proposed approach in Section VIII. Section VIII contains concluding remarks.

\section{Related Work}

Detecting fire is an important issue, modern technology is in dire need of an appropriate detection system that can reduce the damage caused due to a large number of fire accidents taking place everyday \cite{Color1}.  


Initially, the researchers attempted to develop handcrafted techniques for fire detection by focusing on the motion and color properties of the flame detection. One such work done by Thou-Ho et al. \cite{chen2004early} utilized both chromatic and dynamic properties of fire and smoke for the true flame detection. In another work, Celik et al. \cite{ccelik2007fire} tried to distinguish fire from the smoke utilizing two different color spaces and in order to make the classification more robust, they adopted concepts from fuzzy logic to discriminate fire from the other fire like objects. In contrast to  \cite{chen2004early}, where they used RGB color space for flame detection, Turgay  et al. \cite{ccelik2007fire} used the YCbCr color space and made some modifications to overcome the drawback from the previous technique by creating some more generic  rules to detect the fire, but high false detection rate and restriction of detection only at a feeble distance were the associated drawbacks. In addition to the color property, motion has also been taken as the criterion to detect the fire in some works. Rafiee et al. \cite{rafiee2011fire} used the static and dynamic properties of the fire and smoke. However, the false negative rate remains an issue here also due to the presence of other objects in the background with similar color properties as the fire pixels.

Qiu et al. \cite{qiu2012autoadaptive} proposed an auto-adaptive edge detection algorithm for flame detection. In another work, Rinsurongkawong et al. \cite{rinsurongkawong2012fire} used the dynamic properties of the fire for flame detection, but this method also failed with images having pseudo fire like objects in the background. This drawback was overcome by the authors of \cite{mueller2013optical} in which they proposed two optical flow estimators for differentiating fire from the non-fire objects. Mobin et al. \cite{mobin2016intelligent} introduced a fire detection system, Safe from Fire (SFF) that uses multiple sensors to detect fire and smoke distinctly. But the use of multiple sensors, caused the system to be more expensive.

Although these hand-engineered approaches to fire detection are not computationally expensive and can be deployed to economically feasible hardware like Raspberry Pi with a good frame rate, there is the drawback of manual feature extraction from the raw images. This drawback makes the hand-engineering task very tedious and inefficient, particularly when the number of images in the dataset is  high. In contrast, the DL-based approaches have the advantage of automatic feature extraction, thus, making the process much more efficient and reliable than the conventional handcrafted image processing techniques. However, these deep learning approaches require a lot of heavy computational power, not only while training but also when the trained model is to be deployed to hardware for carrying out a specific task. In the case of fire detection, the capability of the algorithm to be deployed on computationally heavy hardware like a personal computer machine is futile because the unit needs to be comparable to a conventional fire detector, both in terms of physical size and cost.

Various deep learning approaches for fire detection have been proposed. Zhang et al.\cite{zhang2016deep} in their work on forest fire detection utilized fire patches detection with a fine-tuned pre-trained CNN, `AlexNet' \cite{krizhevsky2012imagenet} while Sharma et al. \cite{sharma2017deep} also proposed a CNN-based fire detection approach using VGG16 \cite{simonyan2014deep} and Resnet50 \cite{he2016deep} as baseline architectures. But in both these works, the large on-disk size and a number of parameters render these models unsuitable for on-field fire detection applications using  low-cost low-performance hardware. 

Muhammad  et al. fine-tuned different variants of CNNs like AlexNet \cite{muhammad2018early}, SqueezeNet \cite{muhammad2018efficient}, GoogleNet \cite{muhammad2018convolutional}, and MobileNetV2 \cite{muhammad2019efficient}. In  \cite{muhammad2018early,muhammad2018efficient,muhammad2018convolutional}, they used Foggia's dataset \cite{foggia2015real} as the major portion of their train dataset, while in \cite{muhammad2019efficient} the train dataset was combined from \cite{chino2015bowfire} and \cite{foggia2015real}. Although Foggia's dataset contains 14 fire and 17 non-fire videos with a large number of frames, the dataset contains a lot of similar images, which restricts the performance of the model trained on this dataset to a very specific range of images. Moreover, in \cite{muhammad2019efficient} the training dataset consists of 1844 fire images and 6188 non-fire images, which points towards an unbalanced dataset, and consequently the possibility of biased results. More importantly, these networks have a large number of layers and have large on-disk size, which restricts their use in embedded vision application. The incorrect selection of training dataset and bulky models with a large number of layers and parameters direct towards the need for a shallow network, which has a good trade-off between fire detection accuracy and high frame rate, allowing the model to run efficiently on a low-cost embedded device. Therefore, learning from our previous works, we tried to make our train dataset more diverse and challenging by gathering images from Flickr and Google while combining these images with few images sampled from Foggia's and Sharma et al. dataset \cite{sharma2017deep}. To tackle the problem of too big on-disk size and a large number of layers, we build our own neural network from scratch and name it as `FireNet'. Moreover, in order to help other researchers to build better fire detection models, we have open sourced our dataset and trained model. More details about our dataset and network are provided in the subsequent sections.

\section{DATASET DESCRIPTION}

We observed from the datasets used in the past approaches \cite{muhammad2018early,muhammad2018efficient,muhammad2018convolutional,muhammad2019efficient} that currently there is a scarcity of a  diverse fire dataset. One of the dataset provided by Foggia et al. \cite{foggia2015real} contains 31 fire and non-fire videos. Although Foggia's dataset is vast, it is not diverse enough to be solely utilized for training a network and expecting it to perform well in realistic fire detection scenarios. The reason that this dataset does not appear to be diverse enough is that it contains a large number of similar images. Thus, we tried to create a diverse dataset by shooting fire and non-fire videos in a challenging environment, and also by collecting fire and non-fire images with fire like objects in the background from the internet. Our train dataset consists of few fire and non-fire images sampled from the Foggia's and Sharma's \cite{sharma2017deep} dataset, and images taken from the internet (Google and Flickr). In order to maintain the diversity in our train dataset, we also augmented the Sharma's dataset and randomly picked a few images from it. Thus, the final train dataset consists of a total of 1,124 fire images and 1,301 non-fire images. Although the dataset may appear to be small, it is extremely diverse.

For the test dataset, we tried to accumulate as many realistic images as possible because the fire detection unit has to ultimately work in these situations only. However, since our train dataset is already diverse enough, we used these realistic images for testing purpose only.

Our complete test dataset consists of 46 fire videos (19,094 frames) with 16 non-fire videos (6,747 frames) and additional 160 challenging non-fire images. To make our test dataset diverse, out of all the frames extracted from this dataset, we randomly sampled few images from each video to form our final test dataset to be used in this work. 
The model performed extremely well and the results are discussed in section VI. In Fig.~\ref{fig1} and Fig.~\ref{fig2}, we have shown a few images from our training and test dataset respectively.

As a means to let the research community benefit from, and extend our efforts, we have open-sourced the dataset and FireNet \cite{github_ref}.

\begin{figure}[tb]
		\begin{tabular}{lccc}
		{{\includegraphics[width=2.55cm,height=2cm]{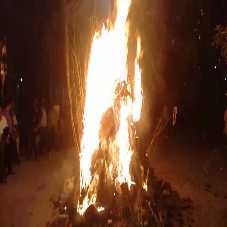}}}&
		{{\includegraphics[width=2.55cm,height=2cm]{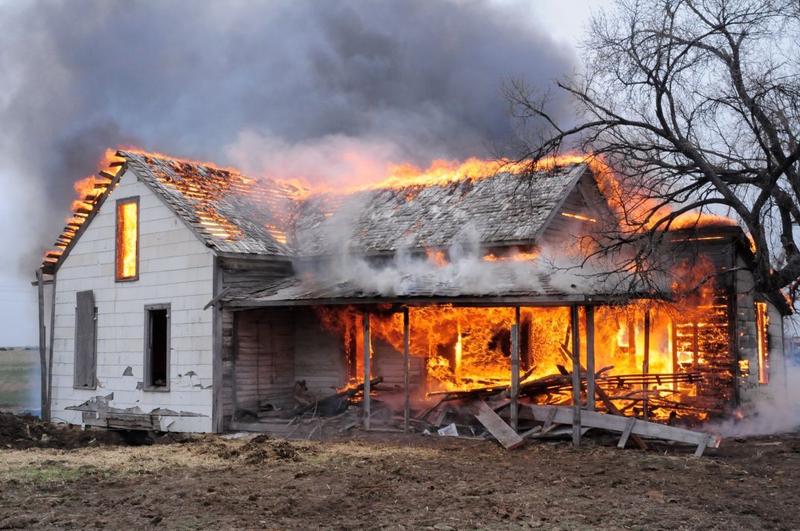}}}&
		{{\includegraphics[width=2.55cm,height=2cm]{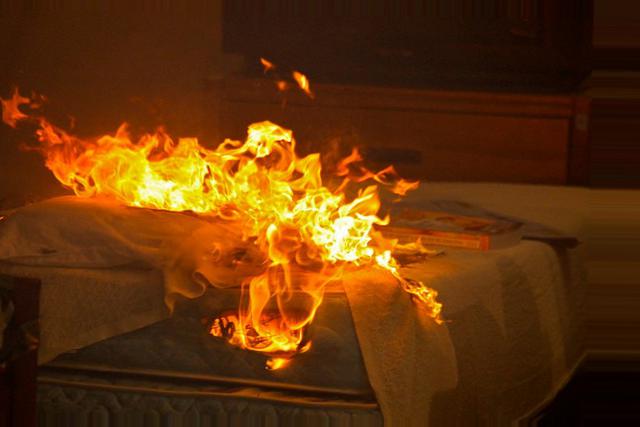}}}\\
		{{\includegraphics[width=2.55cm,height=2cm]{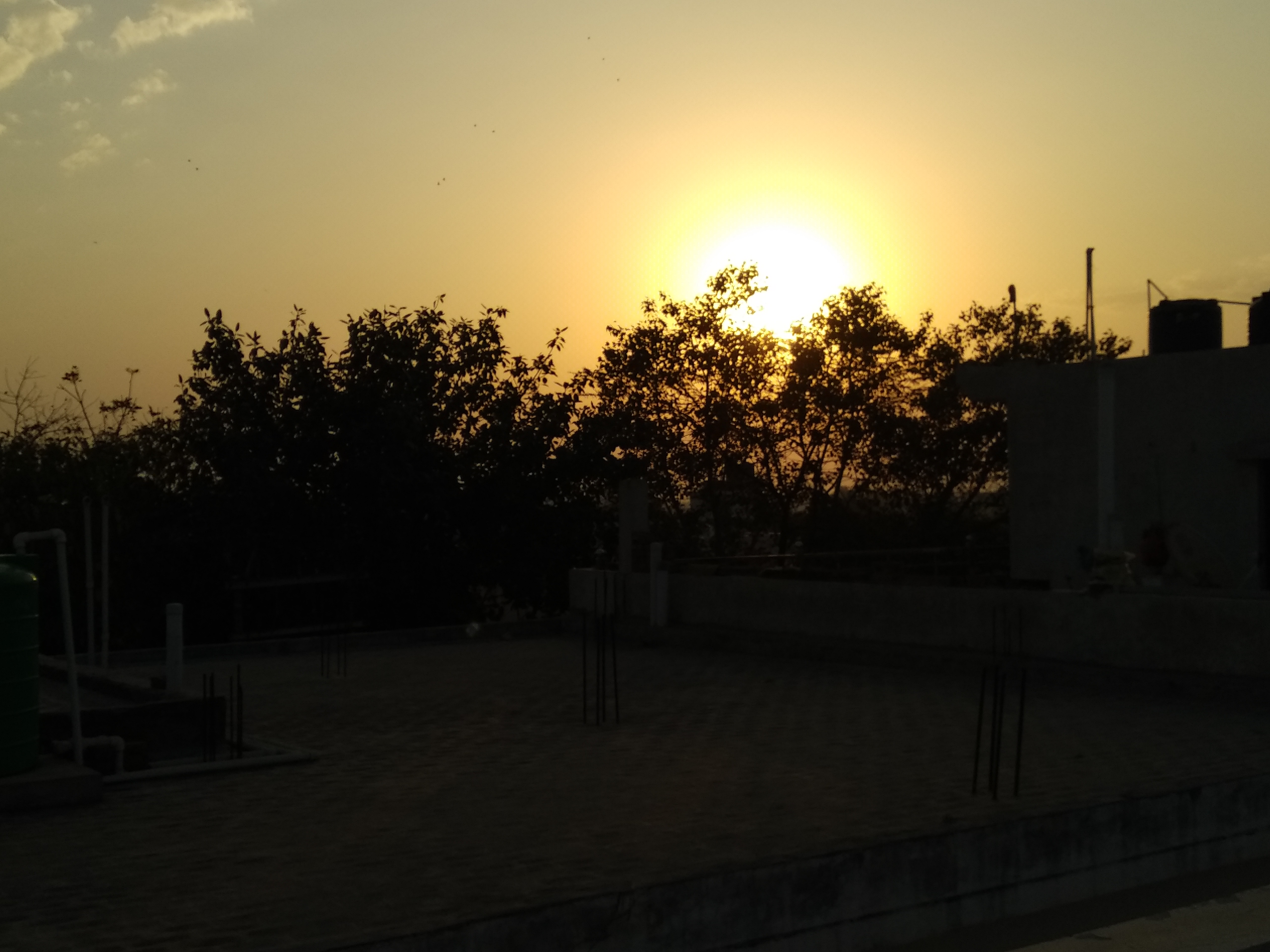}}}&
		{{\includegraphics[width=2.55cm,height=2cm]{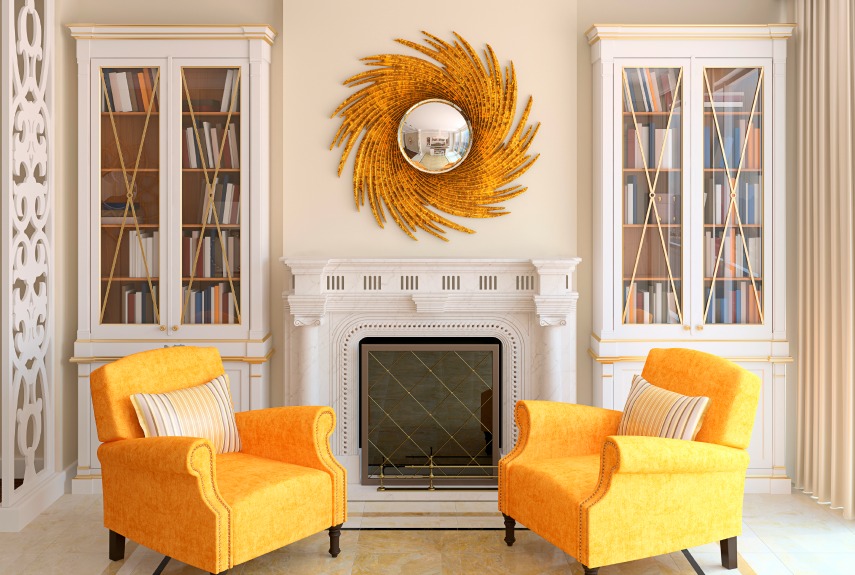}}}&
		{{\includegraphics[width=2.55cm,height=2cm]{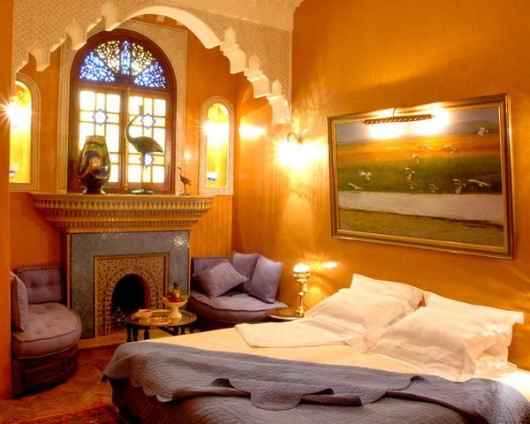}}}\\
		\end{tabular}
	\caption{Few images from our training dataset.}
	\label{fig1}
\end{figure}
\begin{figure}[t]
	\begin{tabular}{lccc}
		{{\includegraphics[width=2.55cm,height=2cm]{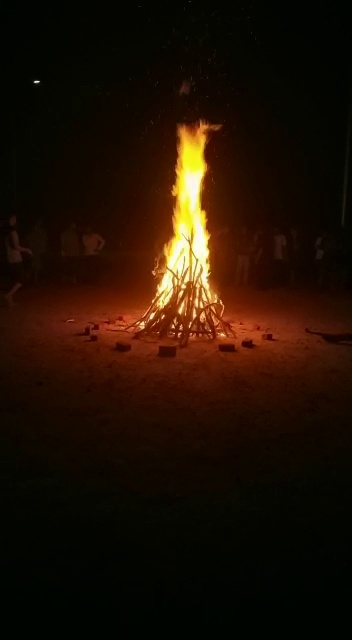}}}&
		{{\includegraphics[width=2.55cm,height=2cm]{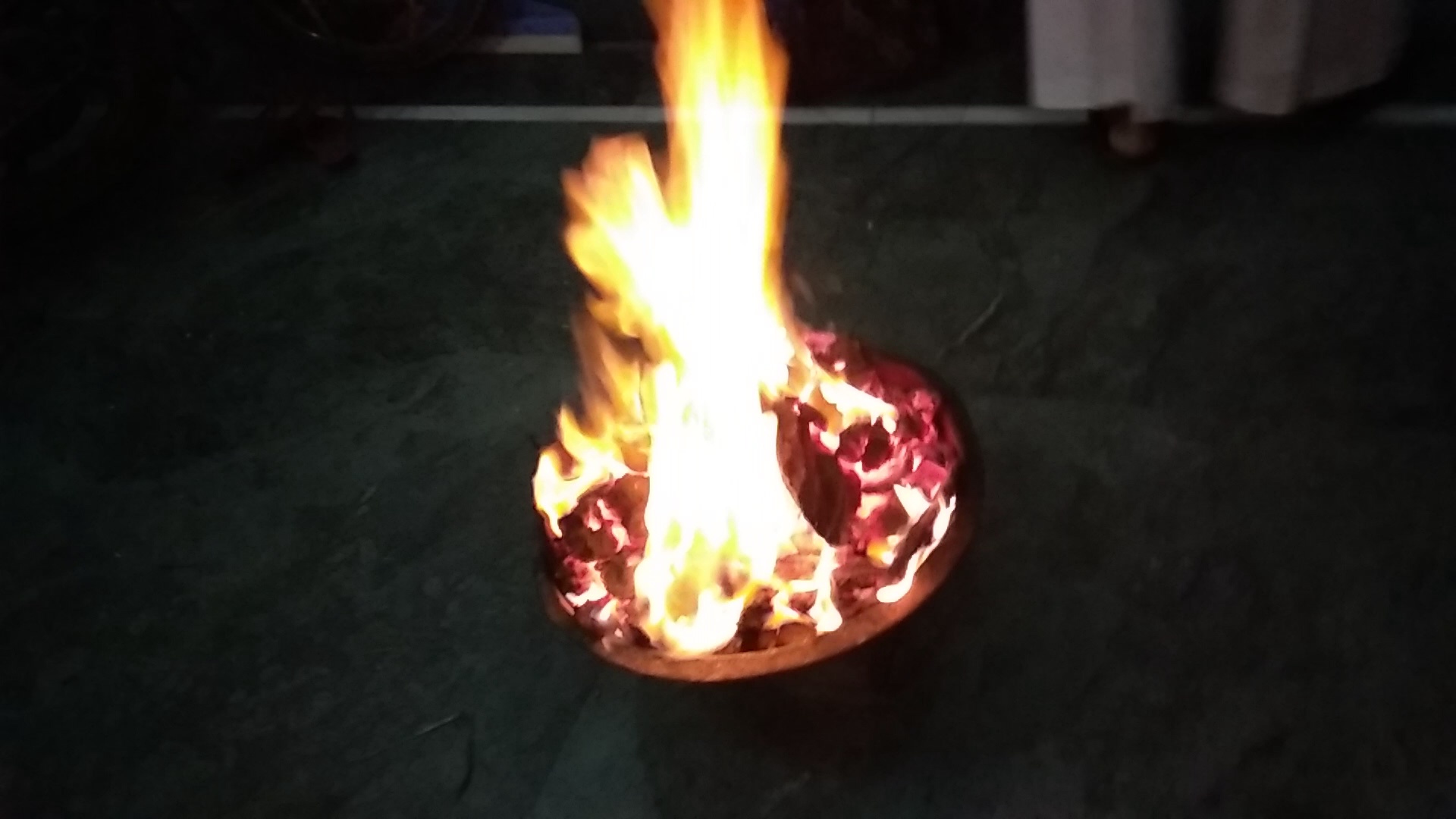}}}&
		{{\includegraphics[width=2.55cm,height=2cm]{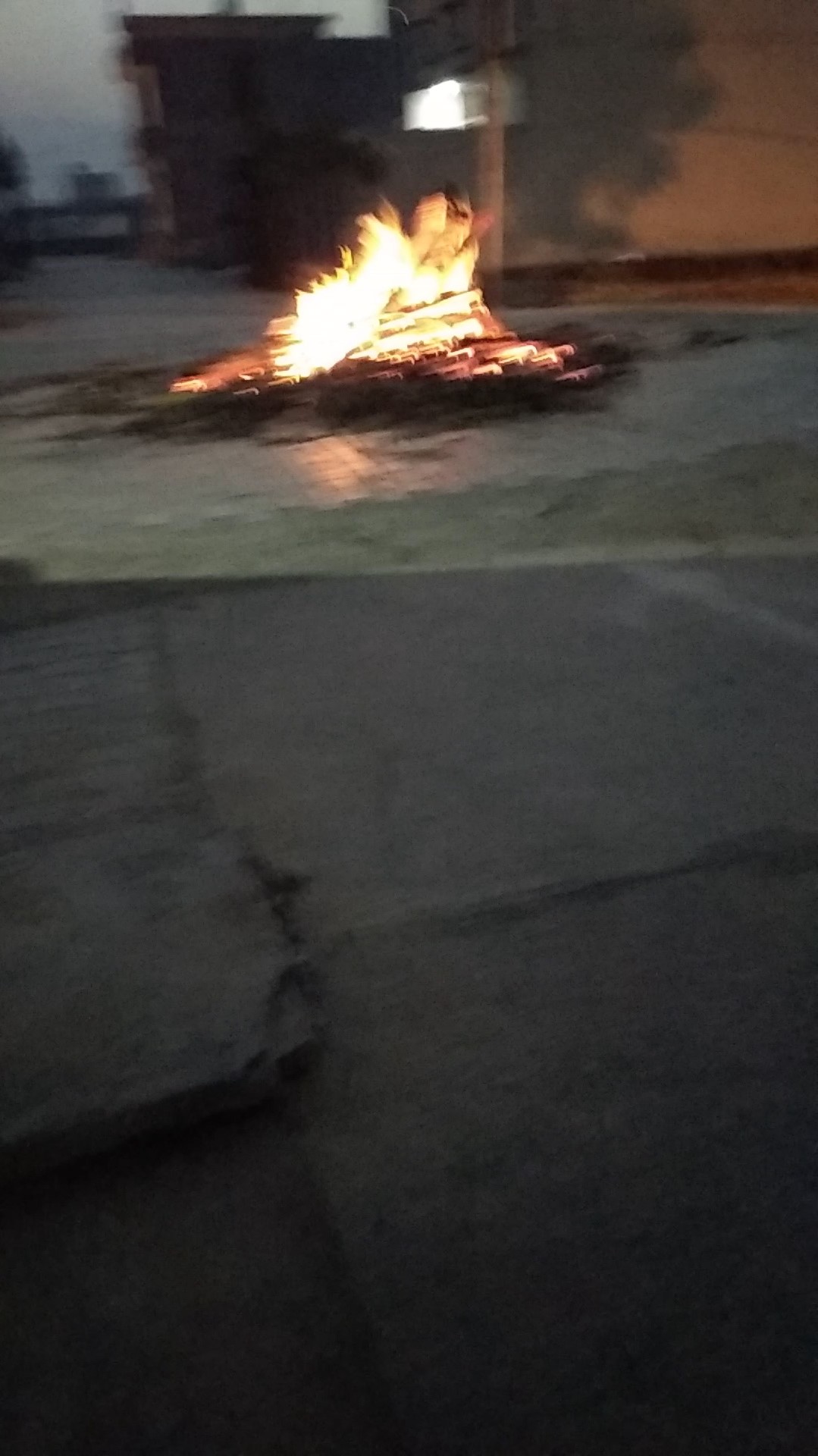}}}\\
		{{\includegraphics[width=2.55cm,height=2cm]{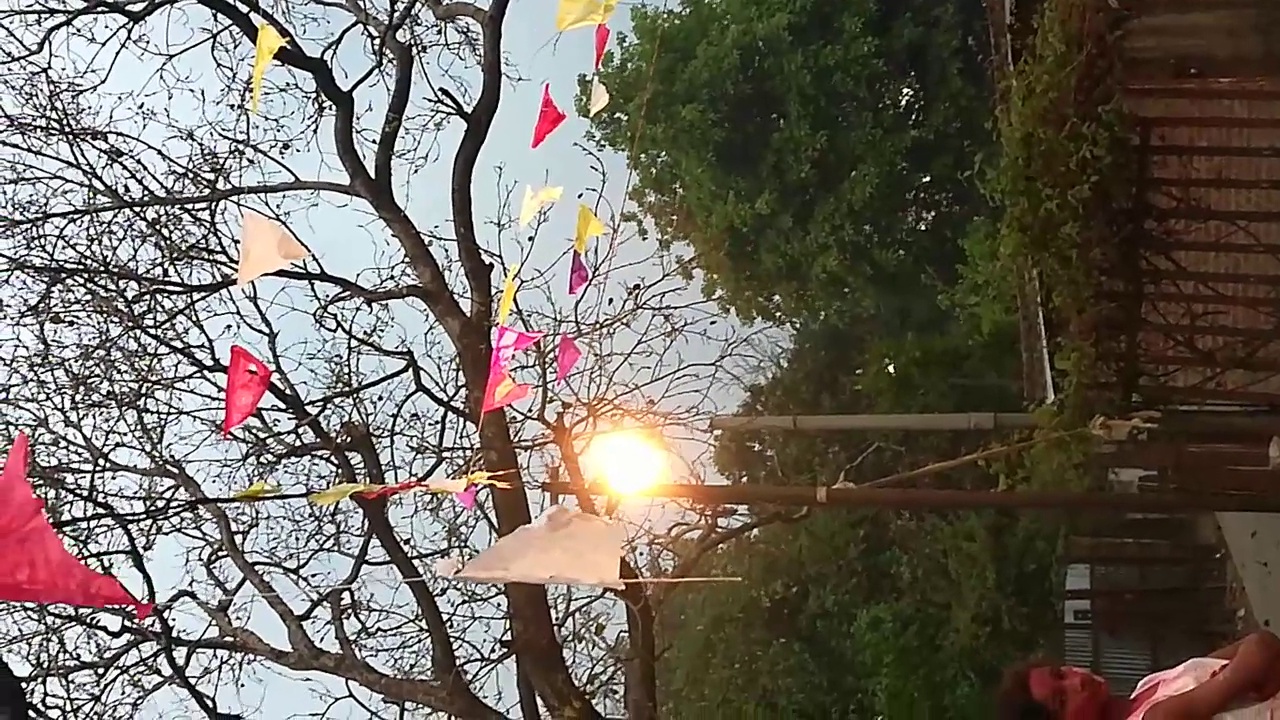}}}&
		{{\includegraphics[width=2.55cm,height=2cm]{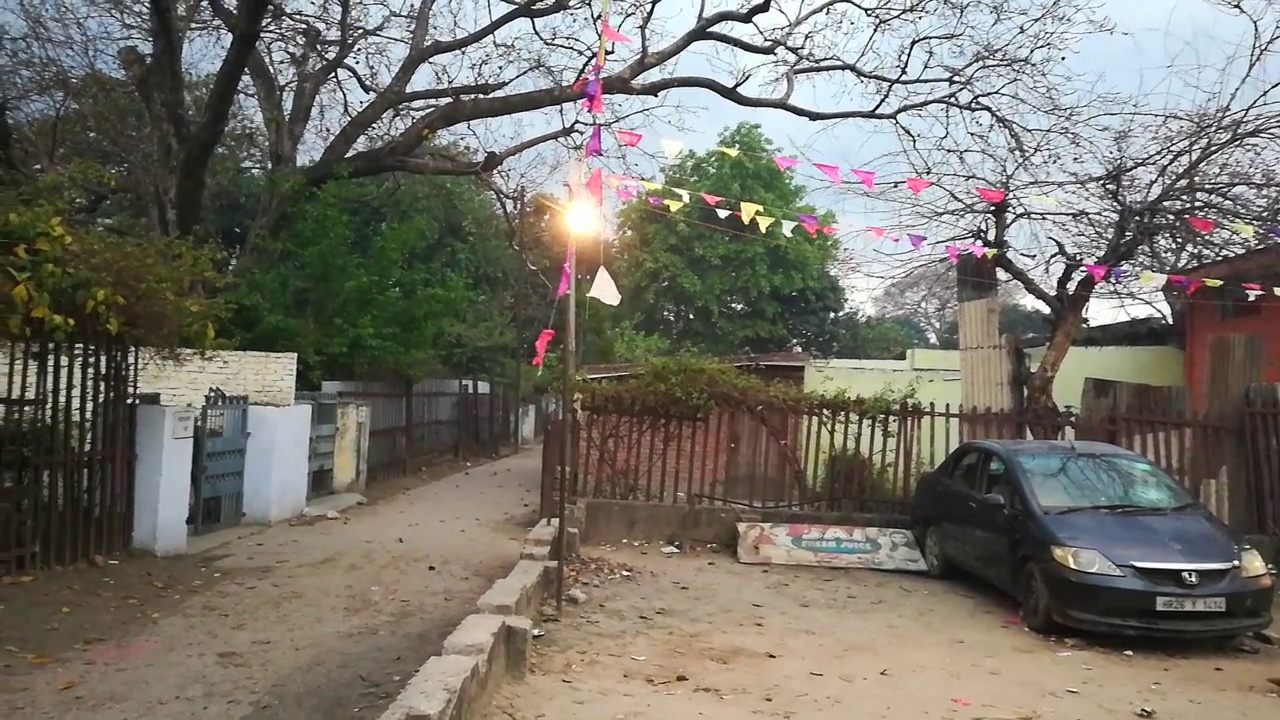}}}&
		{{\includegraphics[width=2.55cm,height=2cm]{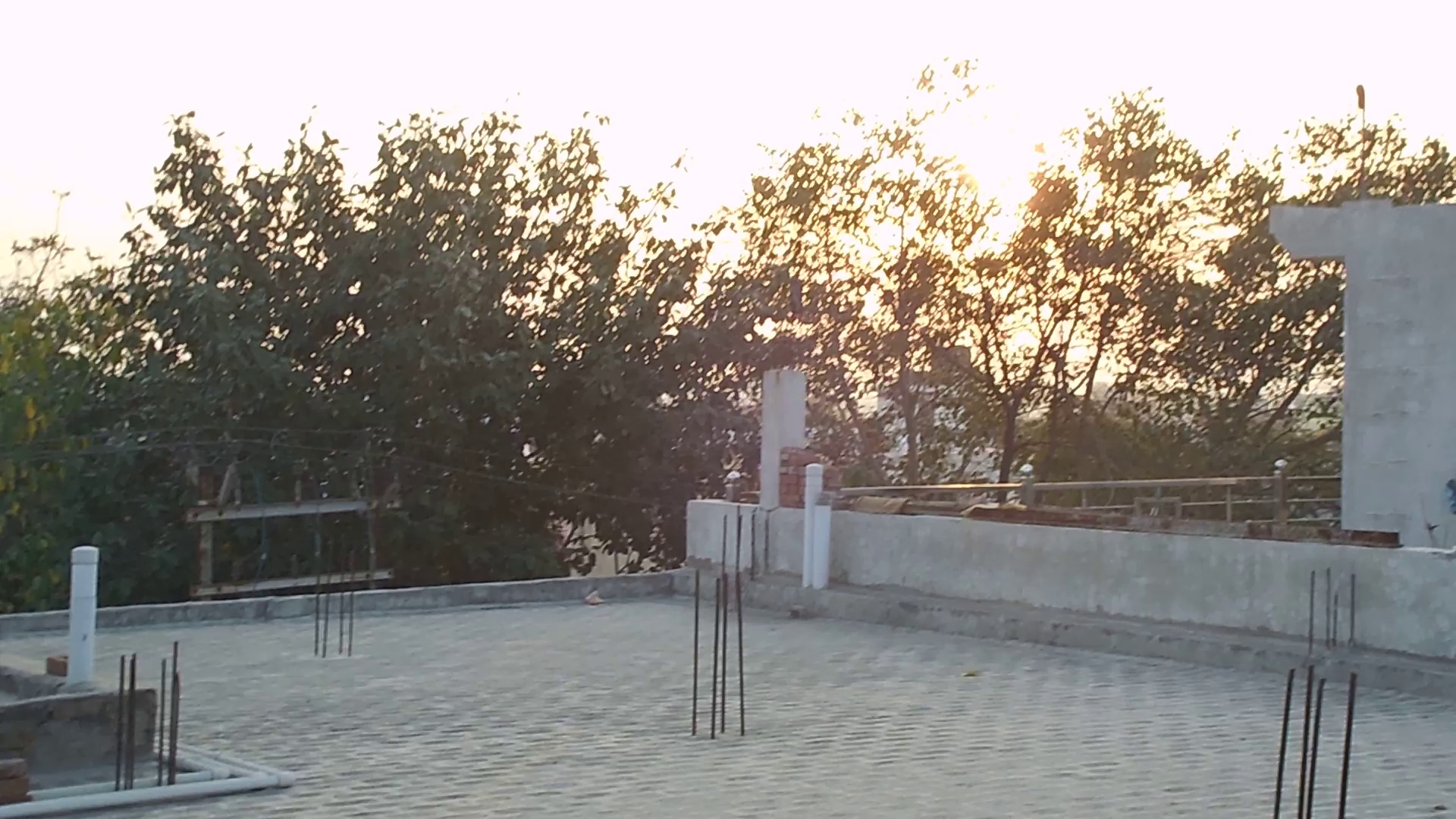}}}\\
	\end{tabular}
	\caption{Few images from our test dataset.}
	\label{fig2}
\end{figure}

\section{PROPOSED APPROACH}

The past deep learning based fire detection approaches like \cite{sharma2017deep,muhammad2018early,muhammad2018efficient,muhammad2018convolutional,muhammad2019efficient} only followed the process of fine-tuning different CNNs like VGG16, Resnet, GoogleNet, SqueezeNet, and MobileNetV2. One major drawback with just fine-tuning such bulky CNNs is that the final on-disk size of the trained model and number of layers is too large, thus preventing these trained models to run smoothly at a sufficient frame rate on low-cost hardware like a Raspberry Pi for real-time fire detection. Also, it is quite obvious that the emphasis of the trained model to run at a good frame rate on low-cost hardware like Raspberry Pi is very much valid, since the end goal of all these approaches is implementation for  real-world applications, i.e., to be transformed into various fire detection units installed in the required environment like a shopping mall, a residence, hotel, etc. Thus, there is the need to use commercially available low-cost hardware, which is economically feasible unlike a high-cost extensive computational machine which is futile in real-world fire detection applications. 

Therefore, we propose a light-weight neural network architecture called FireNet, that is suitable for mobile and embedded applications, which shows a favorable performance for real-time fire detection application. The network runs at a very satisfactory frame rate of more than 24 frames per second on less powerful, economically feasible single board computers like Raspberry Pi 3B, etc. The proposed neural network has three convolution layers and four dense layers (including an output `softmax' layer). 

\subsection{ Architecture}

The complete architecture diagram of the proposed network is shown in Fig.~\ref{fig3}, from where it can be seen that FireNet contains a total of 14 layers (including pooling, dropout and `Softmax' output layer). There are three convolution layers, each of which are coupled with pooling and a dropout layer. Each of these layers has Rectified Linear Unit (Relu) as the activation function except the last layer, which has Softmax as its activation function. The total number of trainable parameters turns out to be 646,818 (size on disk $\sim$7.45 MB). 

The first layer is a convolution layer, which takes a colored input image of size (64$\times$64$\times$3). This input size is selected after empirical results that compared various sizes. The input size can be increased up to (128$\times$128$\times$3) without having a drastic effect on FPS. This layer has 16 filters with a kernel size of (3, 3).

In each of the two subsequent convolution layers, we double the input features keeping the kernel size constant. This is followed by a flatten layer and 2 dense layers of 256 and 128 neurons each. The final layer is a dense layer with two neurons, acting as the output prediction layer. 

\begin{figure}[tb]
\begin{center}
\includegraphics[width=0.48\textwidth]{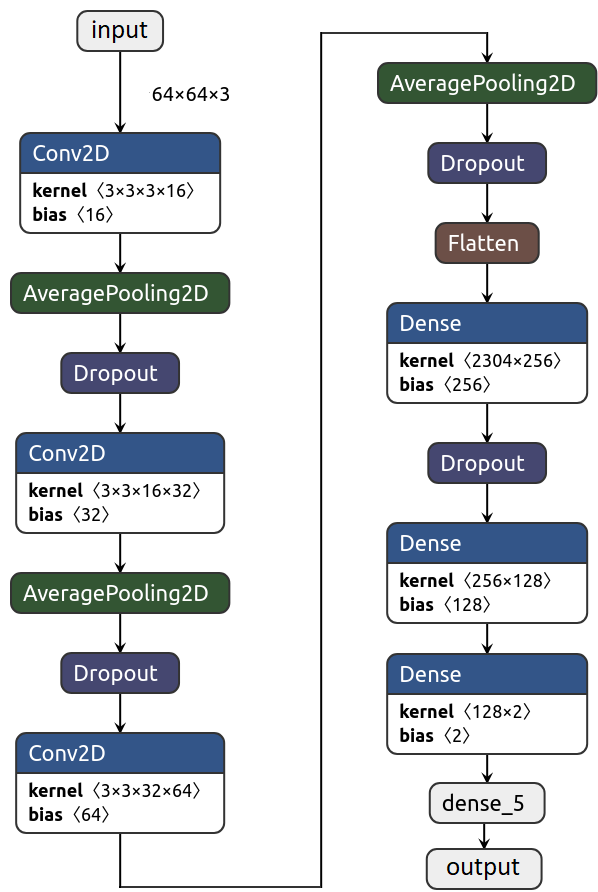}
\caption{Architecture of the proposed neural network.}
\label{fig3}
\end{center}	
\end{figure}

\subsection{Regularization}

We have used dropout with convolution layers along with the dense layers. The general reasoning trend is to use dropout with dense layers only. However, we saw that the overall results of the neural network improved when dropout was used with convolution layers. Hence, we opted for it. We chose a standard dropout value of 0.5 for convolution layers. A dropout value of 0.2 was taken for the subsequent dense layer. This is because it has been shown that overfitting generally takes place in initial layers of the neural  network\cite{srivastava2014dropout}.

\section{Complete Fire Detection Unit and IoT Implementation}

We deployed `FireNet' to the Raspberry Pi 3B and interfaced a smoke sensor and two distinct fire alarms to it, in order to detect fire and smoke distinctly and consequently triggering the distinct fire alarm thus, eliminating the false triggering problem associated with the smoke based conventional fire detectors or alarm systems. We also made the complete fire detection unit IoT capable, thus, allowing the development of a completely autonomous unit with the potential of providing visual fire feedback and alert message in the emergency situation. For implementing IoT based remote visual fire feedback and alert, we utilized two cloud facilities, Twilio \cite{Color3} and Amazon Web Service's Simple Storage Service (AWS S3) \cite{Color4}. Twilio is a messaging service that allows sending SMS/MMS while AWS S3 is a file storage service. We utilize AWS S3 to upload fire images or clips recorded at the time of fire emergency to the cloud while using Twilio to send an MMS containing fire image or clip along with the fire alert message. Fig.~\ref{fig4} shows an overview of the complete IoT enabled fire detection unit capable of differentiating between fire and smoke. The microcontroller has been used as an analog to digital converter (ADC) for communicating the smoke sensor's analog data to Raspberry Pi, as well as to trigger the sound alarms. In a future work, the microcontroller is expected to be done away with, by the inclusion of a special purpose ADC for Raspberry Pi (e.g. MCP3008) in the design. 

\begin{figure}[tb]
	\centering
	{{\includegraphics[width=0.48\textwidth]{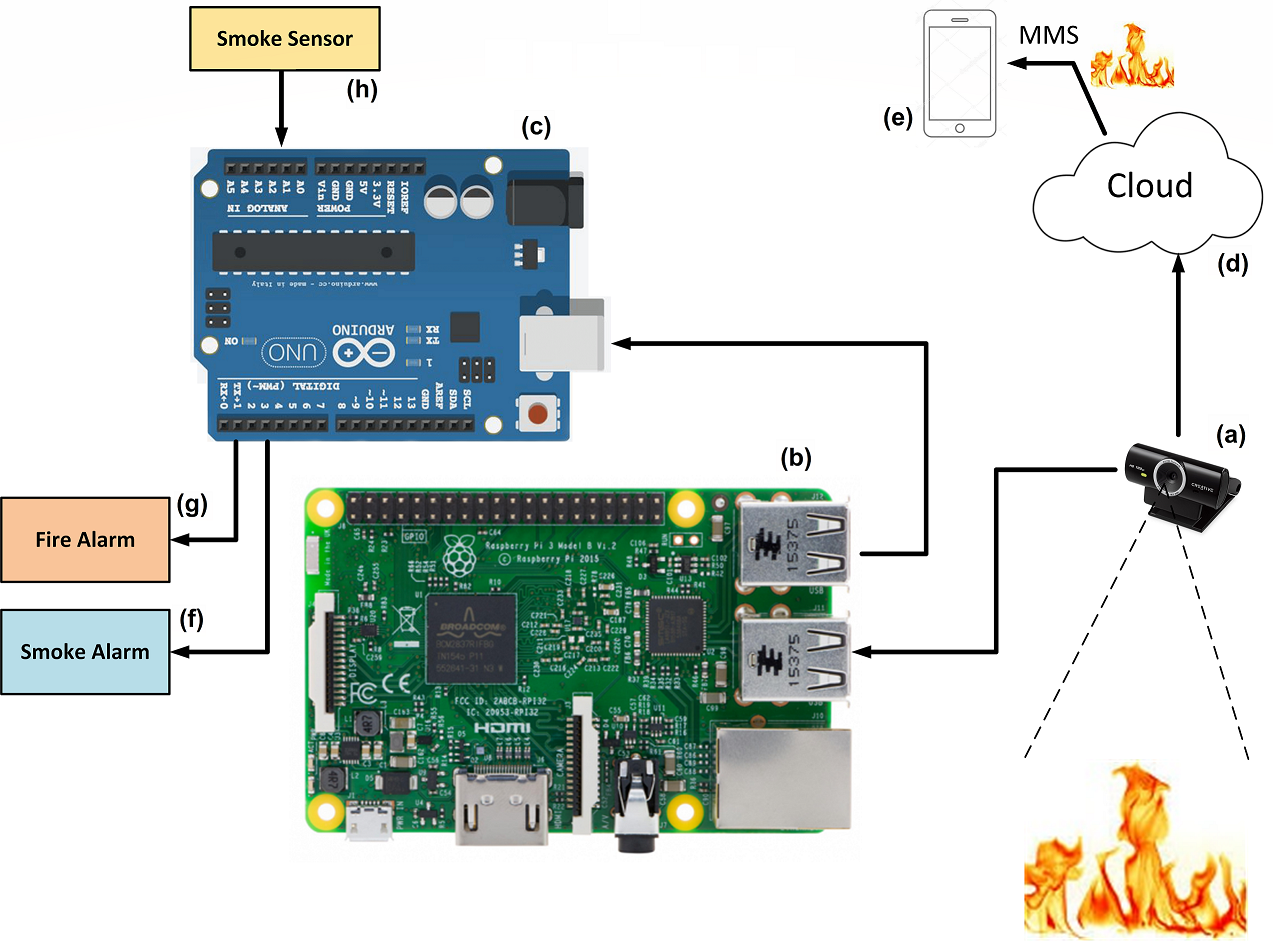}}} 
	\caption{Overview of the complete unit: (a) Camera (b) Raspberry Pi 3B  (c) Microcontroller (d) Cloud storage and SMS/MMS service (Amazon S3 and Twilio) (e) End-user device for receiving fire alert (visual and textual) (f) Smoke alarm (g) Fire alarm with a different sound than smoke alarm (h) Smoke sensor for sensing smoke and thus, aiding in fire-smoke differentiation.}
	\label{fig4}
\end{figure}

\section{Results}

The aim of this work is to present a method that can be smoothly deployed to an embedded device in order to finally build a complete fire detection unit. Therefore, it becomes inevitable to use a test dataset that includes images that are often encountered in real-world fire emergencies with an image quality that is commonly obtained with a camera attached to low-cost hardware like Raspberry Pi 3B. Currently, there is an absence of such a fire dataset. Although the dataset provided in work \cite{foggia2015real} contains such images but the dataset lacks diversity. However, for the sake of comparison we have shown the results of our model on this dataset as well. Therefore, to reflect the performance of our trained model in most realistic situations, we compiled the test dataset from our own videos shot in challenging actual world fire and non-fire environment. The performance obtained on both these datasets is shown in Table~\ref{tab1}. We have used four different metrics (accuracy, precision, recall, and F-measure) in order to present a complete and reliable analysis. As can be seen from Table~\ref{tab1}, the output values obtained for each metric is pretty much up to the mark on both datasets. Moreover, the performance of the model is better on the Foggia's dataset as the image diversity is limited in this dataset. In order to further ensure the reliability of the trained model, we have shown the training process using the training curves. Fig.~\ref{fig5} represents the training and validation loss curves obtained during the training process while Fig.~\ref{fig6} shows the training and validation accuracy curves obtained during the same process. As can be seen from Fig.~\ref{fig5} and Fig.~\ref{fig6}, the trained model is able to generalize significantly on the validation set. Also, note that we have used a 70\% and 30\% split between the training and validation set. Moreover, there is no overlap between the two respective sets. Another significant advantage of this network is that it can run on Raspberry Pi 3B at a frame rate of 24 frames per second.

\begin{table}[tb]
\caption{Test performance of \textquotesingle FireNet\textquotesingle on our real-world test dataset}
\begin{center}
\begin{tabular}{|p{65pt}|p{55pt}|p{55pt}|}
		\hline
		Metrices  & Our dataset (\%) & Foggia's dataset (\%)  \\ \hline
		Accuracy  & \textbf{93.91} & \textbf{96.53}        \\ \hline
		False Positives    & 1.95 & 1.23                    \\ \hline
		False Negatives    & 4.13   & 2.25                 \\ \hline
		Recall    & 94       & 97.46              \\ \hline
		Precision & 97    & 95.54               \\ \hline
		F-measure & 95 & 96.49  \\               \hline 
\end{tabular}
\label{tab1}
\end{center}
\end{table}

\begin{figure}[tb]
\begin{center}
	\includegraphics[width=0.4\textwidth]{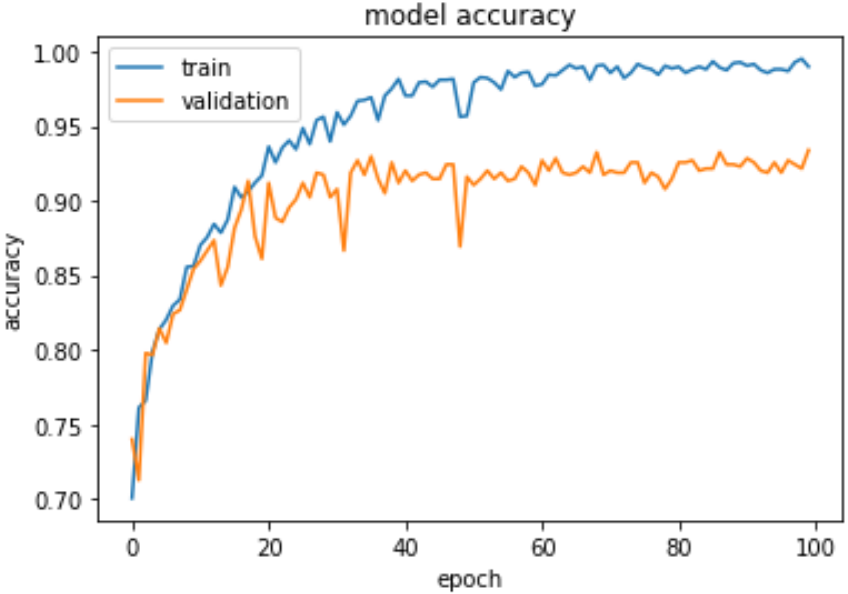}	
	\caption{Training  and validation curves for model accuracy.}
	\label{fig5}
\end{center}
\end{figure}

\begin{figure}[tb]
\begin{center}
	\includegraphics[width=0.4\textwidth]{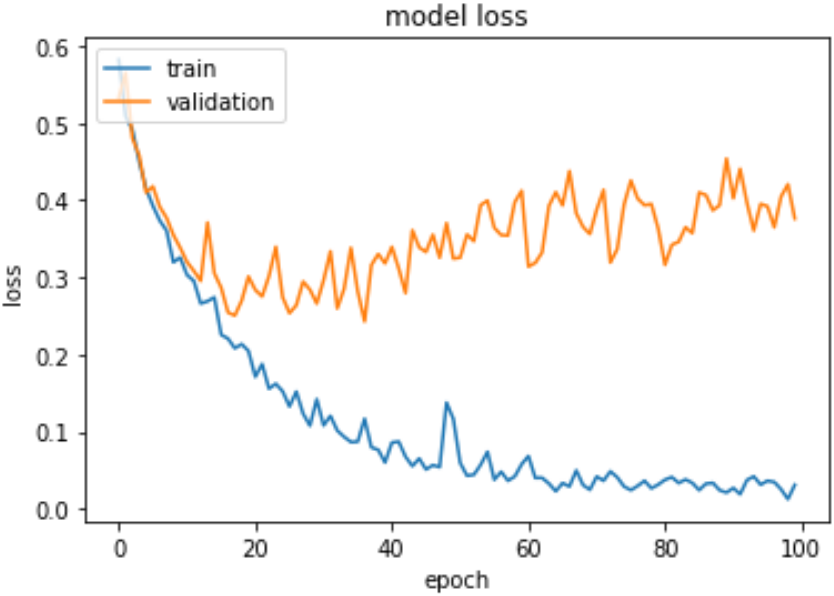}
	\caption{Training and validation curves for model loss.}
	\label{fig6}
\end{center}
\end{figure}

\section{Discussion on similar works}

This section aims to draw a comparison between the proposed FireNet and other available state-of-art approaches for fire detection. The biggest advantage of FireNet is its small size on disk ($\sim$7.45 MB) which is attributed to the shallow network, and  consequently the small number of trainable parameters (646,818). It needs to be acknowledged that there are better performing fire detection solutions available in the literature. However, the definition of `better' is largely dependent upon the resources used and the dataset trained on. For instance, the works in \cite{sharma2017deep,muhammad2018early,muhammad2018efficient,muhammad2018convolutional,muhammad2019efficient} all employed massive CNN based deep models which translate to large on-disk sizes, and typically reported detection capabilities of around 4-5 frames per second while running on low-cost embedded hardware. FireNet on the other hand derives its impressive fire detection capabilities from being trained on a much more diverse dataset, as well as its specialized design from scratch for use in fire detection. Using this effective combination, FireNet is successfully able to provide real-time fire detection feature for upto 24 fps which is almost as good as human visual cognition.

\section{Conclusion}

In this work, we present a very lightweight neural network (`FireNet') build from scratch and trained on a very diverse dataset. The ultimate aim of the complete work is to develop an internet of things (IoT) capable fire detection unit that can effectively replace the current physical sensor based fire detectors and also can reduce the associated problems of false and delayed triggering with such fire detectors. The introduced neural network can smoothly run on a low-cost embedded device like Raspberry Pi 3B at a frame rate of 24 frames per second. The performance obtained by the model on a standard fire dataset and a self-made test dataset (consisting challenging real-world fire and non-fire images with image quality that is similar to the images captured by the camera attached to Raspberry Pi) in terms of accuracy, precision, recall, and F-measure is encouraging. Moreover, the IoT functionality allows the detection unit to provide real-time visual feedback and fire alert in case of fire emergencies to the user. In our future work, we plan to improve the performance of the model on even a more diverse dataset.

\balance
\bibliographystyle{ieeetr}
\bibliography{document}

\end{document}